\documentclass{article}
\usepackage{ijcai26}
\usepackage{times}
\usepackage{url}
\usepackage[hidelinks]{hyperref}
\usepackage[utf8]{inputenc}
\usepackage[small,font=footnotesize,labelfont=bf]{caption}
\usepackage{graphicx}
\usepackage{amsmath}
\usepackage{amssymb}
\usepackage{amsthm}
\usepackage{booktabs}
\usepackage{float}
\usepackage{bbm}

\title{Reading the Room: Implicit Confusion Encoding in Recurrent World Model States}

\author{
Donald Aadithiyan\\
\affiliations
Department of Computer Science and Engineering, University of Moratuwa, Moratuwa, Sri Lanka\\
\emails
donaldaadithiyanwork@gmail.com
}

\begin{document}

\maketitle

\begin{abstract}
World models built on the RSSM architecture, such as DreamerV3, keep a recurrent hidden state $h_t$ trained only to reduce prediction error. We show this state also tracks its own confusion, hiding in plain sight: nearly orthogonal to $h_t$'s directions of greatest variance, invisible to any variance-based method. It is functionally distinct from ensemble disagreement, which flags new inputs, and reconstruction error, which flags bad predictions right now. On a test holding prediction error fixed while confusion varies, a linear probe on $h_t$ finds the signal (AUROC 0.72, 5 runs), while an ensemble baseline scores below chance. A discounted count of recent high-error steps explains 80\% of the probe's output ($R^2=0.80$). We confirm the signal is causally used, not merely present, by editing $h_t$ directly and watching behaviour change, including a check using real values from other trajectories instead of synthetic edits. Its geometry and closed form generalize across three control tasks; the decisive dissociation test itself holds cleanly on only one, and its practical use, deciding when to check reality instead of trusting imagination, generalizes to only two of the three tasks.
\end{abstract}

\section{Introduction}

People often ``read the room'', sensing trouble before anyone says so. This paper asks the same of a model: can it tell, from its own state, that it has been wrong a while, though nothing trained it to track that? DreamerV3 world models keep a recurrent hidden state $h_t$, updated each step by a GRU trained only to reduce reconstruction error and the KL gap between expected and observed. Nothing asks $h_t$ to represent uncertainty, yet representations rarely stay limited to what a loss function asks.

Our question: does $h_t$ encode confusion, and what does it compute? Confused means something precise: surprised for several consecutive steps, not the current one alone, distinct from novelty, a mid-streak model can still get one step right, and a fresh input can be predicted well first try. Ensemble disagreement, the standard model-based-RL tool, cannot tell these apart: it compares predictions across independently trained models, peaking when an input is unfamiliar to all of them, not when one model has struggled quietly on ordinary inputs.

This paper tests three properties of that signal: whether it is separable from novelty and between-model disagreement, not a relabelling; whether the model causally uses it, not merely correlates with it; and whether it is specific to this task and architecture or generalises across structurally different control problems. We test rather than assume, reporting what actually happens, including where the answer complicates a cleaner story.

We build a test separating confusion from novelty directly: hold KL divergence fixed, vary reconstruction quality between two state groups from the same KL band. An ensemble-disagreement baseline scores below chance here, treating novel states as safe and familiar-but-confused states as risky, while a linear probe on $h_t$ scores 0.72: opposite directions on the same test, not two versions of one signal (Table~\ref{tab:setc}).

Standard representation-analysis tools miss this signal too, for a structural reason, not an oversight: PCA finds where most of a representation's variance sits, and the confusion direction sits almost entirely outside that space, 88 degrees from the top 50 components, carrying only 9\% of its own variance there (Section~\ref{sec:results}.3). Any method built around dominant-variance directions would discard it without ever detecting it.

\textbf{Contributions.}
\begin{enumerate}
\item A confusion signal in $h_t$, dissociated from novelty and ensemble disagreement, pointing in opposite directions on one test (Section~\ref{sec:results}.1).
\item A closed-form account of what the probe computes, a discounted count of recent high-error steps, stable across five seeds (Section~\ref{sec:results}.2).
\item Causal evidence the direction is load-bearing, not merely correlated: activation editing against a 50-direction empirical null and an independent real-value-substitution method, replicated across five seeds (Section~\ref{sec:results}.3--4).
\item Generality across three control tasks: the direction, geometry, and closed form all carry over; we trace the exact mechanism behind one metric's inversion on one task (Section~\ref{sec:results}.5).
\item An operational use, deciding when to check a real observation instead of trusting imagination, helping two of three tasks.(Section~\ref{sec:results}.6).
\end{enumerate}

\section{Related Work}

Uncertainty work in RSSM-based world models mostly relies on ensemble disagreement, introduced for exploration by Sekar et al.~\shortcite{sekar2020} and reused widely since \cite{chua2018,kidambi2020,janner2019,lutjens2019,wong2026,bierling2025,hansen2024}: model disagreement, i.e.\ novelty, not accumulated confusion (Section~\ref{sec:results}.1 dissociates the two empirically). Separately, RSSM latents drift toward familiar, high-reward regions off-distribution, masking errors and inflating reward \cite{berger2026}, part of a broader case for physically interpretable world models \cite{peper2025}; our signal provides a complementary inference-time indicator of problem states at no added cost.

Editing an arbitrary subspace can mimic a causal role via an unrelated pathway \cite{makelov2024,sklar2023}, so we adopt the recommended mitigations in Section~\ref{sec:results}.4. Closest prior work uses linear probes to recover task-relevant representations from world models and game-playing agents \cite{nanda2023,taufeeque2024}, probing what the network represents; we instead probe its own predictive confidence. Work probing hidden states for a model's confidence, in language models \cite{azaria2023,burns2023,kadavath2022} or as a trust signal for imagined rollouts in model-based RL \cite{kalweit2017,pan2020,remonda2021,wang2026}, recomputes confidence fresh each decision rather than asking if a persistent summary is encoded. Across all three strands, novelty-detection, subspace-editing validation, and confidence-probing, none distinguish a persistent, accumulated confusion signal from momentary surprise; this paper isolates and causally validates exactly that distinction.

\section{Setup and Method}
\label{sec:setup}

\begin{figure}[H]
\centering
\includegraphics[width=\linewidth,height=2.7in,keepaspectratio]{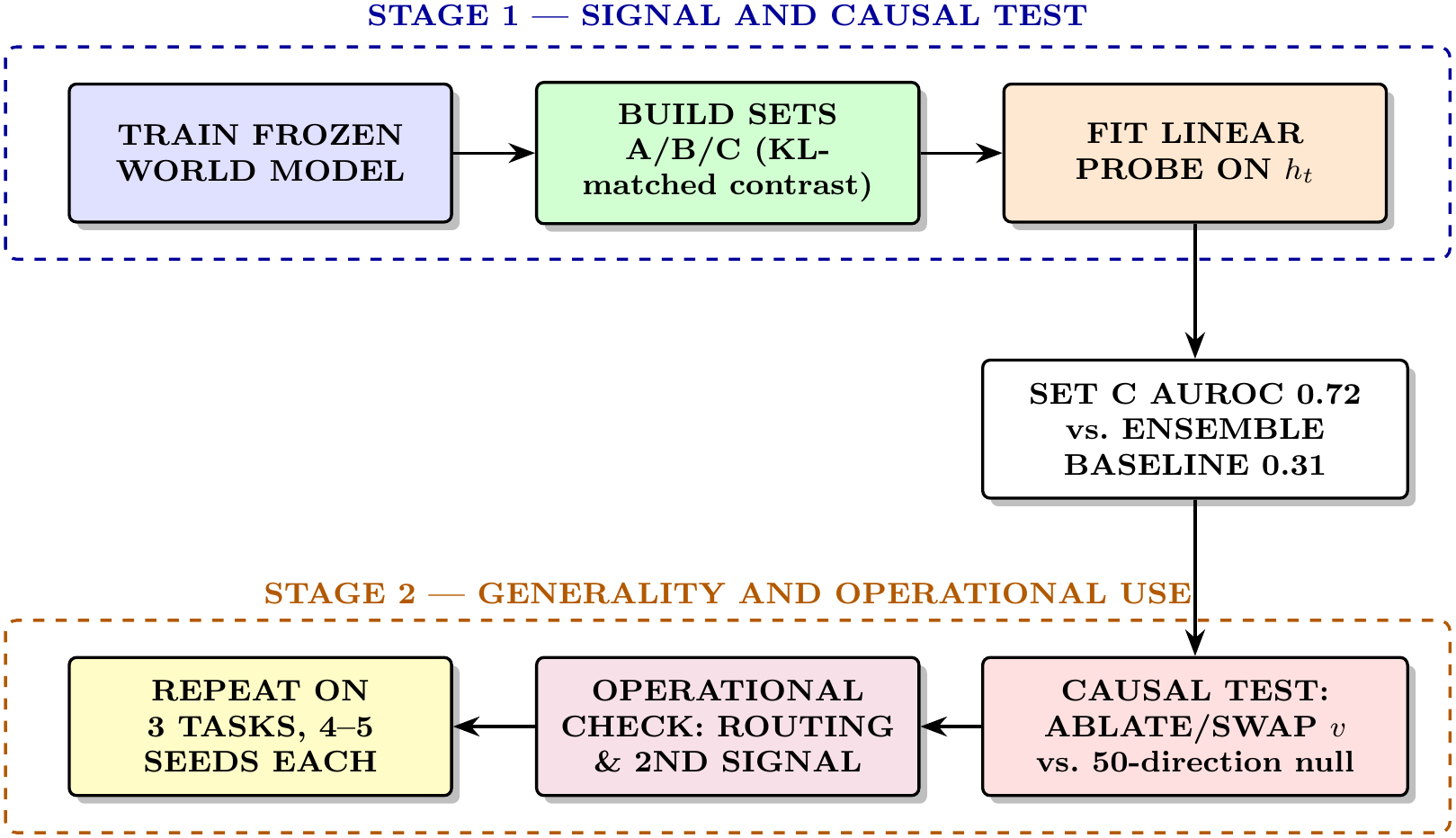}
\caption{The overall pipeline: train once, build the three evaluation sets, fit the probe, test it causally against an empirical null, then repeat across tasks and seeds and check two operational uses.}
\label{fig:pipeline}
\end{figure}

We train a Mini-DreamerV3 (XS configuration: 256-dimensional GRU state, $32\times32$ categorical latents, about 12 million parameters) from scratch for 100,000 steps on dm-control, cartpole-swingup, reacher-easy, and pendulum-swingup. We record $(h_t, z_t, \mathrm{KL}_t, \mathrm{recon}_t)$ throughout training and build three evaluation sets from a frozen model:
\begin{itemize}
\item \textbf{Set A}: fresh in-distribution states.
\item \textbf{Set B}: the same states with Gaussian observation noise.
\item \textbf{Set C}, the decisive test: A and B pooled, then split by reconstruction error within matched KL bins, so the two groups share a KL distribution but differ roughly $9\times$ in reconstruction error, any probe scoring above chance reads something beyond KL magnitude.
\end{itemize}

Full construction, including the noise-leakage and cross-task controls, is in Appendix~\ref{app:setup}. All probes are logistic regression on $h_t$ predicting whether KL is above the median, evaluated by AUROC, reported with a bootstrap 95\% CI or a mean and standard deviation across seeds.

We characterise the probe's output with a discounted count of recent high-KL steps,
\begin{equation}
C_t = \sum_i \gamma^i \,\mathbbm{1}[\mathrm{KL}_{t-i} > \mathrm{median}],
\label{eq:ct}
\end{equation}
where $\mathrm{KL}_{t-i}$ is the model's KL divergence $i$ steps before $t$; $\mathbbm{1}[\cdot]$ is the indicator function (1 if the condition holds, 0 otherwise), so $\mathbbm{1}[\mathrm{KL}_{t-i}>\mathrm{median}]$ marks a high-KL ("confused") step; $\gamma\in(0,1)$ is a decay rate down-weighting older steps; and the sum runs over past steps $i=0,1,2,\dots$, so $C_t$ is a running, decayed count of recent high-KL steps. We fit $\gamma$ by regressing probe score on $C_t$ across a grid of candidate values.

\textbf{Proposition.} If $h_t$ is a sufficient statistic for the model's own recent prediction-error history under a fixed decay rate, a probe trained on binarised KL labels and one trained directly on $C_t$ should reach approximately the same Set C score, since both targets are then linear functions of the same quantity. We test this directly in Section~\ref{sec:results}.2 rather than assume it, since the two probes need not agree in practice even if the proposition holds in principle.

To test causal relevance, we edit $h_t$ on a frozen model at inference time: subtracting the confusion direction, adding a scaled multiple of it, and, as an independent check that avoids synthetic edits entirely, substituting the confusion-direction component from a different real held-out state (full matching procedure in Appendix~\ref{app:causal}). Writing $\mathrm{probe}(\cdot)$ for the trained probe's output, $v$ for the unit confusion direction, and $h_t' = h_t - (h_t\!\cdot\!v)v$ for the ablated state (i.e.\ $h_t$ with its component along $v$ removed), the effect at look-ahead $k$ is
\begin{equation}
\Delta_k = \mathrm{probe}(h_{t+k} \mid h_t') - \mathrm{probe}(h_{t+k} \mid h_t),
\label{eq:delta}
\end{equation}
where $\mathrm{probe}(h_{t+k}\mid h_t)$ is the probe's confusion score $k$ steps after starting from the original state $h_t$, and $\mathrm{probe}(h_{t+k}\mid h_t')$ is the same quantity starting from the ablated state $h_t'$; $\Delta_k$ is the difference the edit makes to the later readout ($k=0$: immediate effect; $k>0$: whether it persists). $\Delta_k$ is compared against the same quantity for 50 random unit directions of matched scale in place of $v$. Every causal claim in Section~\ref{sec:results}.4 is a comparison of $\Delta_k$ against this empirical null, not a single control, following the diagnostic in Sklar~\shortcite{sklar2023}. Full protocol in Appendix~\ref{app:causal}.

\section{Results and Discussion}
\label{sec:results}

\subsection{Confusion Is Not Novelty}

Table~\ref{tab:setc} reports AUROC on each evaluation set. On Set A, the fresh in-distribution baseline, the probe and ensemble score almost identically (0.863 versus 0.868), no meaningful difference between them yet. On Set C, the probe scores 0.72 (5-seed mean $0.715\pm0.074$, 95\% CI $[0.666,0.763]$), while a 5-model ensemble baseline, on the identical construction, scores 0.550, near chance (Appendix~\ref{app:setup}). On a direct novelty-detection test with no KL matching, the pattern reverses: reconstruction error alone reaches 0.996 untrained, the ensemble also does well here (0.966), while the $h_t$ probe drops to 0.49. Confusion and novelty are dissociated in both directions: all three methods occupy different points on the ROC plane, none a weaker version of another. ROC curves and the full Set A/B/C comparison are in Appendix~\ref{app:setup}, Figures~\ref{fig:roc_setA}--\ref{fig:roc_setC}.

\begin{table}[H]
\centering
\small
\begin{tabular}{lccc}
\toprule
Method & Set A & Set C & Direct novelty \\
\midrule
Probe on $h_t$ & 0.863 & \textbf{0.723} & 0.49 \\
Ensemble (5 models) & 0.868 & 0.550 & \textbf{0.966} \\
Reconstruction error & --- & --- & \textbf{0.996} \\
\bottomrule
\end{tabular}
\caption{Confusion and novelty are dissociated. Ensemble and probe are scored on the same Set C construction throughout.}
\label{tab:setc}
\end{table}

\subsection{A Closed-Form Account, Tested Rather Than Assumed}

Regressing probe score on the discounted count in Equation~\ref{eq:ct} gives $R^2=0.798$ at $\gamma=0.95$, identical to two decimal places across all 5 seeds; current KL alone explains only $R^2=0.519$. The gap between these two numbers is the point: the probe is reading accumulated history, not just how confused the model is on the current step alone. A $\gamma$ of 0.95 is roughly a 13-step memory. A second probe trained directly on continuous $C_t$, rather than binarised KL labels, reaches almost the same Set C score (0.711 versus 0.714), consistent with the proposition above, though this closeness is an empirical finding, not a restatement of it, since the two probes could in principle have diverged.

\subsection{Geometry, and Why It Is Not a Gate Artefact}

The confusion direction sits at 88.2 degrees from all of the top 50 principal components of $h_t$, carrying only 9\% of its own variance inside that subspace. This is why the signal is invisible to standard probing: any method that looks where a representation's variance is concentrated would search the wrong part of $h_t$ entirely and never find it. Writing $u_1,\dots,u_{50}$ for the top 50 principal components of $h_t$ (the 50 directions of greatest variance in the hidden state, found by PCA), $U=\mathrm{span}(u_1,\dots,u_{50})$ for the subspace they define, and $v$ for the unit confusion direction, we measure the angle between $v$ and its own projection onto $U$,
\begin{equation}
\theta = \arccos\left(\|\mathrm{proj}_{U}(v)\|/\|v\|\right),
\label{eq:angle}
\end{equation}
where $\mathrm{proj}_U(v)$ is the projection of $v$ onto $U$, its closest point inside the top-50-variance subspace; $\|\cdot\|$ is vector length, so $\|\mathrm{proj}_U(v)\|/\|v\|$ is the fraction of $v$'s length surviving projection onto $U$ (0 to 1); and $\arccos$ turns that fraction into an angle. $\theta$ near $90^\circ$ means almost none of $v$ survives, i.e.\ $v$ lies almost entirely outside $U$.

A geometry this extreme invites two easy but wrong explanations, and we rule out each with a direct test rather than take either on faith:
\begin{itemize}
\item \textbf{Maybe it's just a saturated gate.} A GRU update gate pinned near 0 or 1 would overwrite most of the previous state on every step, which could plausibly push unrelated content into whatever low-variance corner is left over, an artefact of the gate's behaviour rather than a real, separately-encoded signal. We test this directly: forcing the gate to any fixed value between 0.5 and 0.99 at inference time leaves $\theta$ near $90^\circ$ throughout, moving by only 1.29 degrees across the entire range. Gate saturation does not create this geometry.
\item \textbf{Maybe it's just one region of $h_t$.} If a single block of units happened to carry the signal, orthogonality would be a coincidence of where that block sits, not a genuine property of the representation as a whole. Splitting $h_t$ into four equal blocks, the signal is close to equally readable from each (AUROC 0.87 to 0.90 per block), so no single localised region is responsible either.
\end{itemize}
Both explanations fail, leaving the geometry itself, not an artefact of training dynamics or layout, as the more likely account.

\subsection{The Direction Is Causally Load-Bearing}

\begin{table}[H]
\centering
\small
\begin{tabular}{lcc}
\toprule
Method & $\Delta$ probe score & Null percentile \\
\midrule
Ablation (synthetic edit) & $-0.586$ & 100th ($z\approx-22$) \\
Real-value substitution & $-0.761$ & 100th \\
\bottomrule
\end{tabular}
\caption{Two structurally different causal edits agree.}
\label{tab:causal}
\end{table}

If the confusion direction only correlated with confusion by coincidence, $\Delta_0$ (the edit's immediate effect) should look like the null. Instead, subtracting it out of $h_t$ changes the probe score by $-0.586$, the most extreme point (100th percentile, $z\approx-22$) of the 50-direction null, and an independent method that never edits $h_t$ synthetically, splicing in the confusion-direction value from a different real state, reproduces the same result at $-0.761$, again the 100th percentile (Table~\ref{tab:causal}). The effect degrades gradually rather than collapsing when $v$ is rotated slightly, and its decay across look-ahead tracks the same $\gamma=0.95$ time constant the probe reads (Appendix~\ref{app:causal}), consistent with a genuine effect. Replication across 5 separate from-scratch training runs is likewise uneven by design, not by accident: the probe-decay effect replicates on all 5, the routing-decision effect on 3 of 5, and next-step prediction error does not clearly change on any (full breakdown in Appendix~\ref{app:causal}). The direction is what the model reads its own confusion from, but not the mechanism behind its short-term prediction accuracy; these are separate properties.

\subsection{Generality Across Three Tasks, and Why One Metric Inverts}

\begin{table}[H]
\centering
\small
\begin{tabular}{lccc}
\toprule
 & Cartpole & Reacher & Pendulum \\
\midrule
Geometry angle & $88.0^\circ$ & $89.4^\circ$ & $88.1^\circ$ \\
$C_t$ $R^2$ & 0.80 & 0.26 & \textbf{0.86} \\
Set C AUROC & 0.72 & 0.62--0.67 & \textbf{0.32--0.40} \\
Within-bin correlation & $+0.39$ & $-0.09$ & $\mathbf{-0.12}$ \\
\bottomrule
\end{tabular}
\caption{Cross-task comparison. Reacher and pendulum entries are ranges across 4 seeds each.}
\label{tab:threeenv}
\end{table}

Table~\ref{tab:threeenv} compares cartpole, reacher, and pendulum. Geometry carries over closely ($88.0$--$89.4^\circ$ across all three), and so does the closed-form encoding, though unevenly ($C_t$ $R^2$ 0.80 cartpole, 0.86 pendulum, only 0.26 reacher). The exact memory length and Set C do not: pendulum's Set C score inverts to 0.32--0.40 despite the strongest $C_t$ fit. The table's last row explains why: the correlation between reconstruction error and $C_t$ computed within the matched KL bins is positive on cartpole ($+0.39$) but negative on reacher ($-0.09$) and pendulum ($-0.12$), predicting each Set C result's direction (Appendix~\ref{app:crossenv}). This is a practical check any user of a reconstruction-based contrastive test can run before trusting it on a new environment: the sign of this one correlation, not just the resulting AUROC.

\subsection{An Operational Use: Deciding When to Check Reality}

\begin{figure}[H]
\centering
\includegraphics[width=\linewidth,height=2in,keepaspectratio]{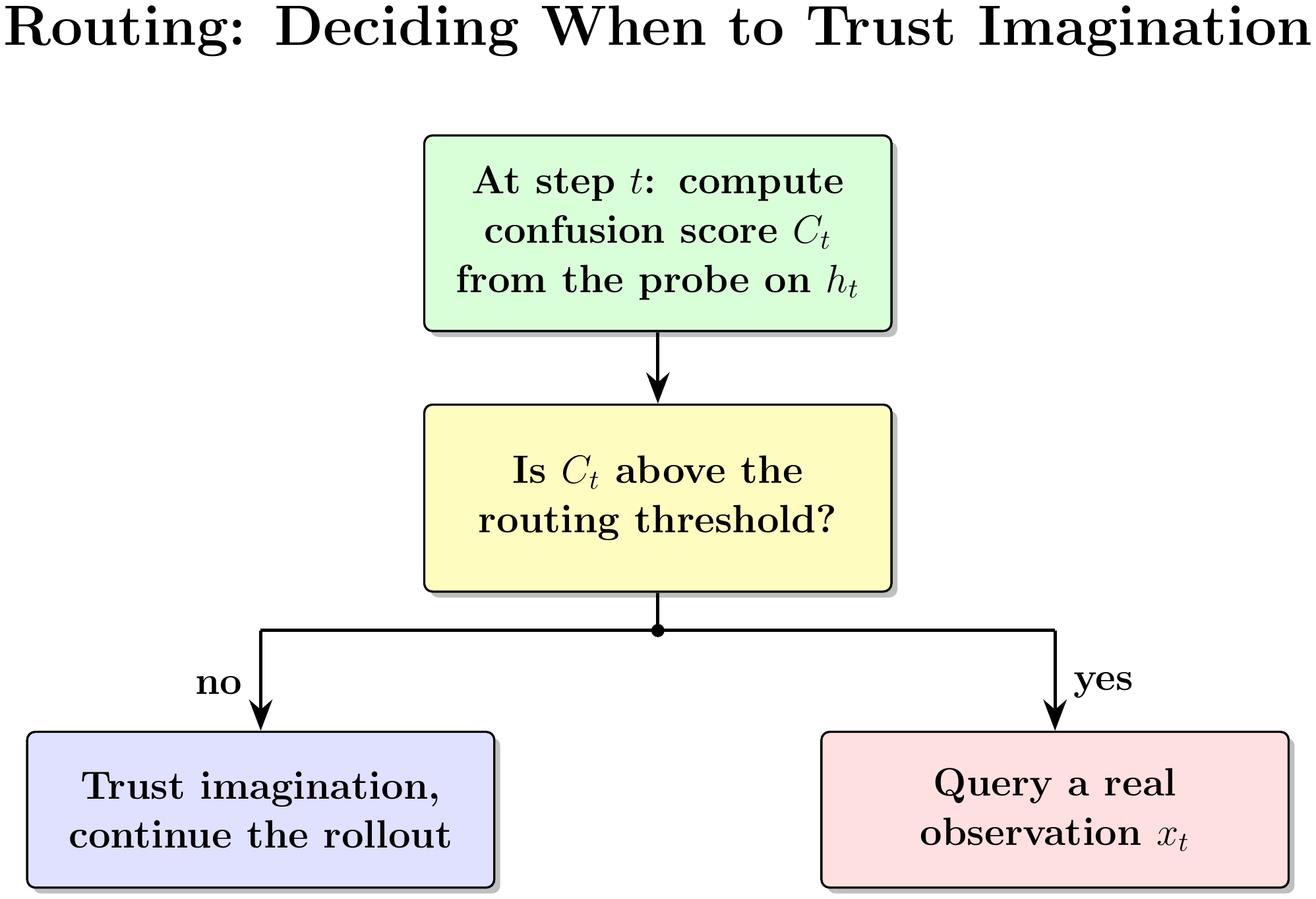}
\caption{The routing decision made at every step: threshold the probe's confusion score to decide whether to query a real observation or continue trusting imagination. Resulting recall against a reconstruction-error baseline, per task, is in Table~\ref{tab:routing}.}
\label{fig:routing}
\end{figure}

Using the probe score to decide when to query a real observation beats a reconstruction-error threshold on cartpole (0.818 vs.\ 0.770 recall) and pendulum, and loses on reacher (Table~\ref{tab:routing}). Pendulum's gain holds despite Set C inverting there, since routing uses KL events directly, bypassing Set C; reacher is the one loss, where reconstruction error is already strong on its own. Against a raw-KL threshold, the probe wins only on reacher ($+0.273\pm0.020$, 3/3 seeds), shows no difference on cartpole ($-0.006\pm0.025$, unstable), and is worse than KL on pendulum ($-0.030$, 3/3 seeds; Appendix~\ref{app:routing}): reacher, where $C_t$'s fit is weakest (Section~\ref{sec:results}.5), is where confusion is least redundant with KL, and where the probe earns its keep.

\begin{table}[H]
\centering
\small
\begin{tabular}{lc}
\toprule
Task & Recall gain vs.\ recon-error baseline \\
\midrule
Cartpole & $+0.04$ \\
Pendulum & $\mathbf{+0.30}$ \\
Reacher & $-0.08$ \\
\bottomrule
\end{tabular}
\caption{Routing recall gain at a 30\% query budget.}
\label{tab:routing}
\end{table}

 A $C_t$-direct router tracks Probe-A closely rather than rescuing reacher or pendulum (Appendix~\ref{app:routing}), robust to the Set C artefact but not reacher's dynamics.

\section{Limitations}
\label{sec:limitations}

Four limitations qualify these results. First, all experiments use a 256-dimensional GRU, three orders of magnitude below full DreamerV3 (4096-dimensional, roughly 200 million parameters); the geometry and causal-editing results hold at double the width (Appendix~\ref{app:scale}), one data point toward that gap, scaling to full size is the natural next step. Second, reacher remains the one consistent exception across three checks, traceable to its own reconstruction error already being an unusually strong, confusion-aligned signal. Third, two attempts to directly correct value estimates from imagined rollouts, down-weighting and early stopping, both failed cleanly (Appendix~\ref{app:negative}): the signal detects confusion, not yet corrects for it. Fourth, routing beat reconstruction error on two of three tasks; against KL alone, the probe's advantage is only on reacher, and should be checked on new tasks (Appendix~\ref{app:routing}).

\section{Conclusion}

\textbf{This work aims to characterise an emergent property of a small, controlled world model, not to claim readiness for safety-critical deployment or generalisation to large-scale systems.} A DreamerV3 world model's hidden state learns to track its own recent confusion, distinct from novelty and ensemble disagreement, without being asked to during training. The signal has a simple closed-form description, is causally load-bearing rather than merely correlated, and carries over to structurally different control tasks, including a case where a specific evaluation metric misleadingly inverts for a traceable, checkable reason. It is directly usable to decide when an agent should stop trusting its own imagination. A world model, it turns out, can tell when it is out of its depth, and act on it, without ever being taught the words for it.

Every result here ran on a single laptop CPU, proof that causally-validated interpretability findings do not need frontier-scale compute, lowering the barrier for researchers without large clusters. Routing is a direct efficiency payoff: reserving costly real observations only when needed, useful for agents in resource-scarce settings. We hope this encourages more interpretability and safety research to be attempted, and trusted, at small scale.

\appendix

\section{Full Experimental Setup}
\label{app:setup}

\textbf{Model.} Mini-DreamerV3, XS configuration: 256-dimensional deterministic GRU state $h_t$, $32\times32$ categorical stochastic state $z_t$, about 12 million parameters, matching the smallest configuration in the original DreamerV3 ablations. The GRU input each step is $[z_{t-1}, a_{t-1}]$, so $h_t = \mathrm{GRU}([z_{t-1},a_{t-1}], h_{t-1})$; stochastic sampling uses straight-through categorical, the one source of run-to-run non-determinism (\S\ref{app:causal}). Only GRU and MLP width change between XS and the full XL model (4096-dimensional, over 200 million parameters); stochastic state size is unchanged, why the geometry finding is worth a scale check (Appendix~\ref{app:scale}) rather than assumed size-specific.

\textbf{Training.} Selected hyperparameters (\texttt{XS\_CONFIG}):

\begin{table}[H]
\centering
\small
\begin{tabular}{ll}
\toprule
Hyperparameter & Value \\
\midrule
Env steps per model & 100{,}000 \\
Sequence length / batch size & 16 / 8 \\
Learning rate / grad clip & 3e-4 / 1.0 \\
Replay capacity & 600 episodes \\
Episode max steps & 500 \\
OOD noise $\sigma$ (Set B) & 0.1 \\
\bottomrule
\end{tabular}
\caption{Selected training hyperparameters; full list in the companion supplement.}
\label{tab:hparams}
\end{table}

Analysis collection uses a uniform-random policy on the frozen model, queried but never updated. All runs used CPU rather than GPU/MPS, faster and more reproducible at this batch size.

\textbf{Tasks.} dm-control, cartpole-swingup (5-dimensional observation, 1-dimensional action), reacher-easy (6-dimensional, 2-dimensional), pendulum-swingup (3-dimensional, 1-dimensional). 100,000 training steps each.

\textbf{Evaluation sets.} All three sets are built from a single frozen, trained world model, no further training involved:
\begin{itemize}
\item \textbf{Set A (in-distribution).} 10,000 fresh states collected by running the frozen model on the task it was trained on, no perturbation.
\item \textbf{Set B (near out-of-distribution).} The same 10,000 states as Set A, but with $\sigma=0.1$ Gaussian noise added to the observation before it reaches the model.
\item \textbf{Set C (the decisive KL-matched contrastive test).} Pool Set A and Set B, split into 10 bins by KL percentile, then within each bin take the bottom 30\% by reconstruction error as group 1 and the top 30\% as group 2: matched KL by construction (22.9 vs.\ 23.9 nats) but $9\times$ different in reconstruction error (0.052 vs.\ 0.471), so any probe separating the groups reads something beyond KL magnitude.
\end{itemize}

The same three-set construction and probe-training protocol described here is applied independently and identically to each of the three tasks, producing the per-task numbers in Table~\ref{tab:threeenv} and Appendix~\ref{app:crossenv}.

\begin{figure}[H]
\centering
\includegraphics[width=\linewidth,height=2in,keepaspectratio]{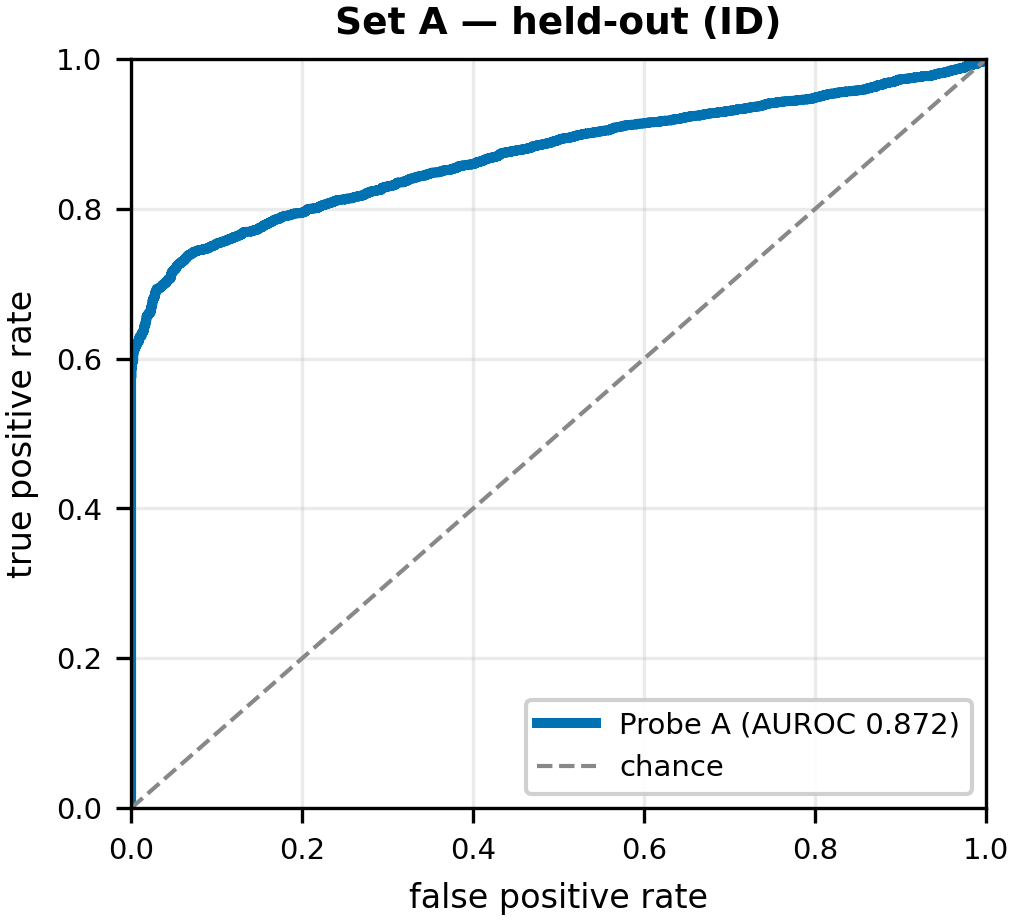}
\caption{Set A (held-out, in-distribution) ROC curve, AUROC 0.872.}
\label{fig:roc_setA}
\end{figure}

\begin{figure}[H]
\centering
\includegraphics[width=\linewidth,height=2in,keepaspectratio]{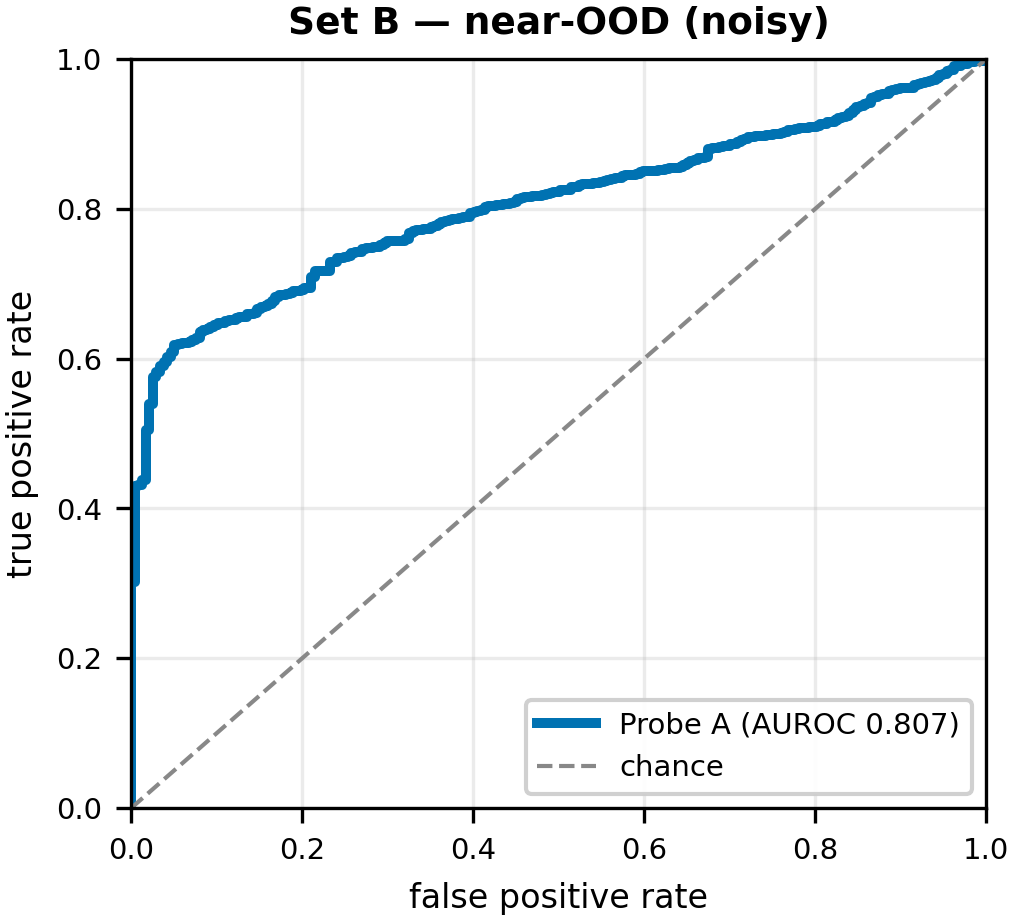}
\caption{Set B (near-OOD, Gaussian-noised) ROC curve, AUROC 0.807.}
\label{fig:roc_setB}
\end{figure}

\begin{figure}[H]
\centering
\includegraphics[width=\linewidth,height=2in,keepaspectratio]{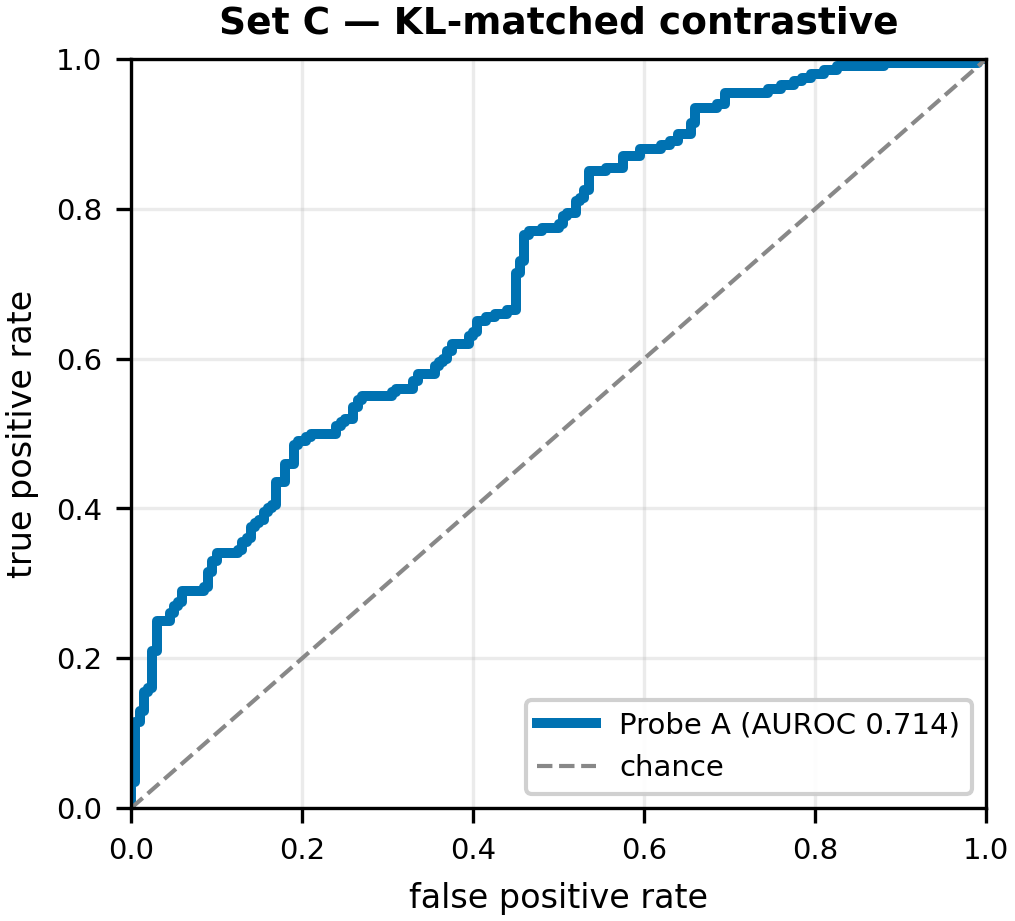}
\caption{Set C (the KL-matched contrastive test) ROC curve, AUROC 0.714.}
\label{fig:roc_setC}
\end{figure}

As a check against noise leakage between Set A and Set B, we rebuilt Set C using only Set A states: AUROC 0.7115 versus 0.7144 for the pooled version, smaller than measurement noise. A separate control set, built the same way from an untrained task (cartpole-balance), tests cross-task transfer directly.

\textbf{Probe training.} Logistic regression (scikit-learn, L2 penalty, $C=1$, lbfgs, max 2000 iterations, random\_state=0) with standard scaling fit on the training split only, stratified 60/40 train/test split. This makes the probe deterministic given a fixed $h_t$ matrix; the only non-determinism entering the ROC figures is upstream $z$-sampling during collection (Appendix~\ref{app:causal}). The 5-seed replication on cartpole, and 4-seed replication on reacher and pendulum, each retrain the full pipeline independently and report mean and standard deviation with bootstrap 95\% confidence intervals (1000 resamples) on the headline numbers. Best $\gamma$ is selected by ridge regression of $C_t$ onto $h_t$ on the test split, and is stable across all 5 cartpole seeds.

\begin{figure}[H]
\centering
\includegraphics[width=0.85\linewidth,height=3in,keepaspectratio]{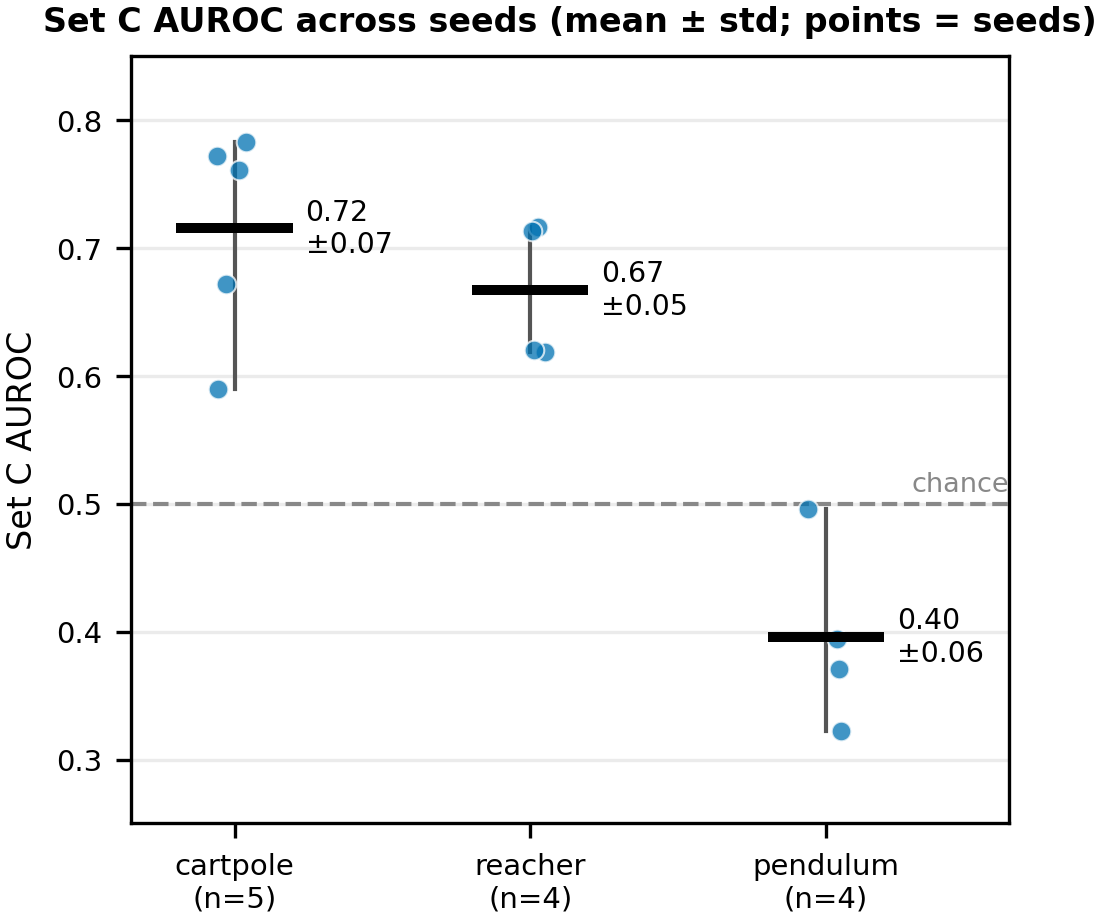}
\caption{Set C AUROC per seed (points) with mean and range (bar and whisker), for each of the three tasks. Individual seed values are shown rather than only a bar, since a bar with error bars can visually imply more precision than 4--5 points actually support. Pendulum's inversion below chance is visible directly, not just in the mean.}
\label{fig:cross_seed}
\end{figure}

\textbf{Ensemble baseline.} A 5-model ensemble-disagreement score (two additional members trained at the identical XS configuration, seeds 3--4, bringing the total to 5, matching the probe's own 5-seed replication count), computed as the variance across independently trained world models' predicted rewards or observations during imagined rollouts, the standard approach for epistemic uncertainty elsewhere in model-based reinforcement learning. Sweeping ensemble size from 2 to 5 members shows the Set C gap does not narrow with a larger, better-resourced ensemble (0.561 at $n=2$ vs.\ 0.550 at $n=5$), while the same larger ensemble's performance on direct novelty detection climbs monotonically (0.867 to 0.966): the ensemble is a genuinely strong novelty detector, and it still cannot separate the KL-matched Set C groups.

\section{Full Causal Validation}
\label{app:causal}

This section gives the full protocol behind Section~\ref{sec:results}.3--4's causal claims, building on the probing and causal-editing setup introduced in Section~\ref{sec:setup} and diagrammed below in Figure~\ref{fig:probe_causal}. All causal edits use a pool of 600 held-out states, disjoint from Probe A's training split, so generalisation is satisfied by construction; the confusion direction is consistent across sub-runs (cosine 0.78).

\begin{figure}[H]
\centering
\includegraphics[width=\linewidth,height=3in]{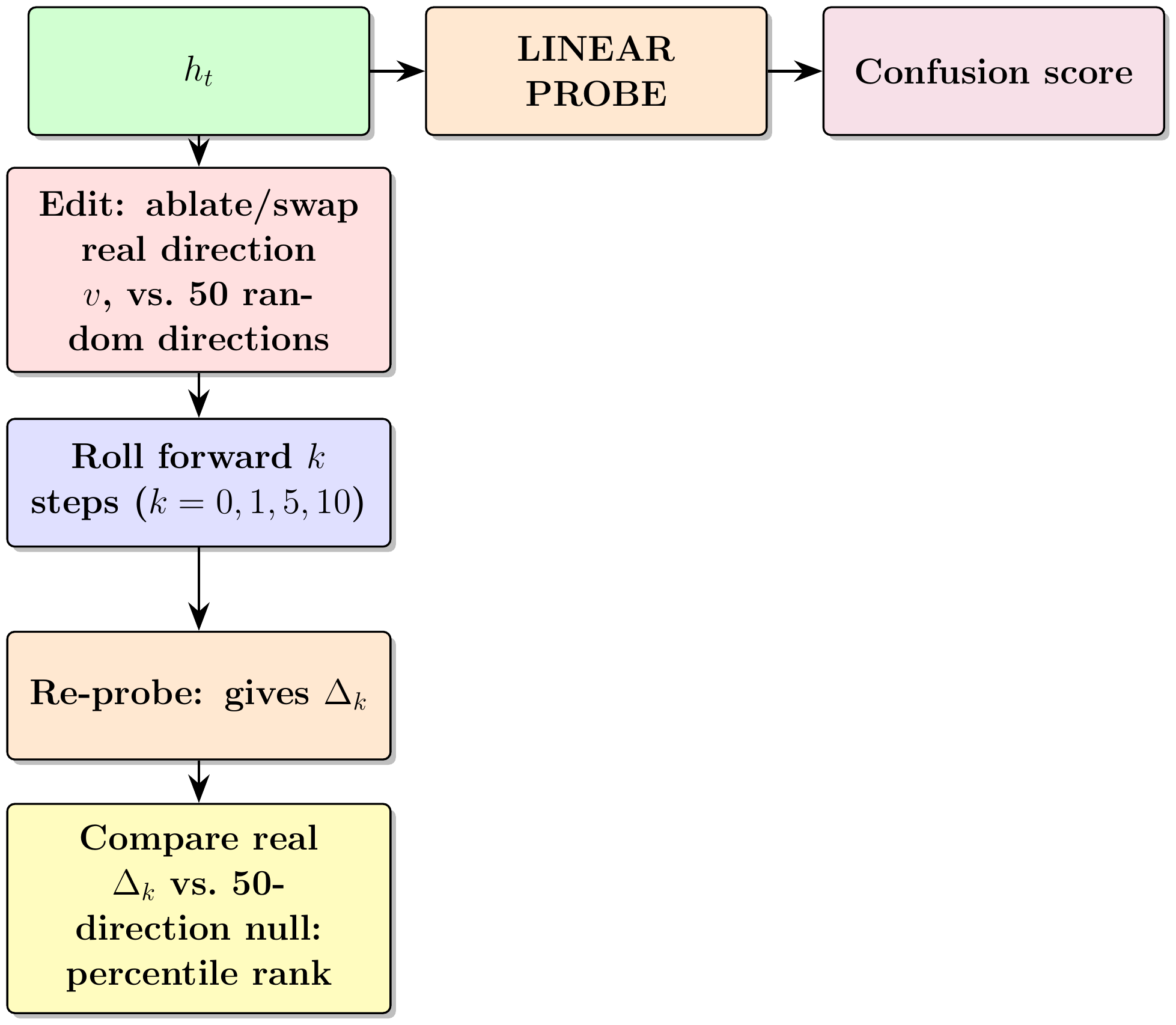}
\caption{The probing and causal-test setup summarised (Section~\ref{sec:results}.3).}
\label{fig:probe_causal}
\end{figure}

\begin{figure}[H]
\centering
\includegraphics[width=\linewidth,height=2in,keepaspectratio]{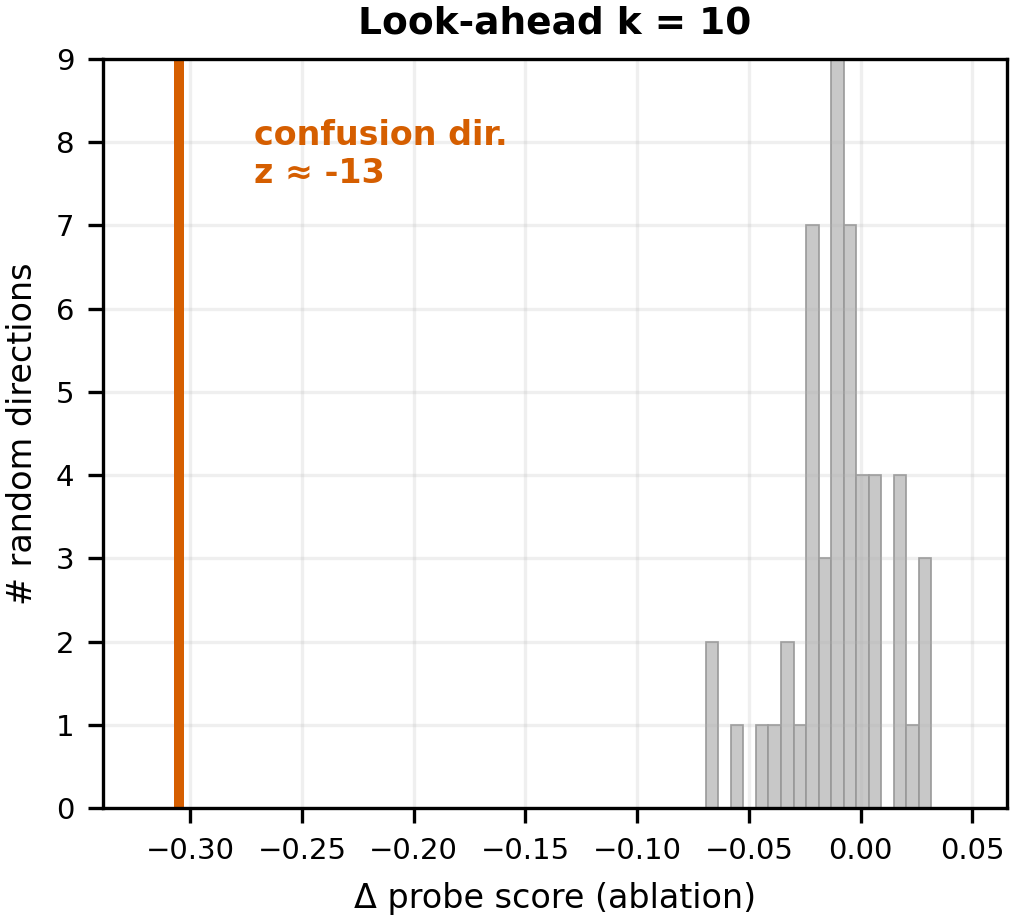}
\caption{Ablation effect against the 50-direction empirical null at look-ahead $k=10$. The confusion direction (marked) sits at the 100th percentile of the null, $z\approx-13$, showing the effect persists but decays with $k$.}
\label{fig:null_k10}
\end{figure}

\textbf{Empirical null.} On 600 held-out states not used to train the probe, subtracting the confusion direction changes the probe score by $-0.586$ at the point of the edit, against a null of $0.001\pm0.026$ from 50 unrelated random directions of matched scale, the 100th percentile of the null throughout ($z\approx-22$). The same pattern holds at later look-ahead steps, 100th percentile at every $k$ ($z\approx-19,-15,-13$ at $k=1,5,10$; Figure~\ref{fig:null_k10}).

\textbf{The decay tracks $\gamma$.} The closed-form model of Section~\ref{sec:results}.2 predicts that removing step $t$'s contribution should reduce the downstream ablation effect by $\gamma^k$.

\begin{table}[H]
\centering
\small
\begin{tabular}{lcc}
\toprule
Look-ahead $k$ & Observed $|\Delta_k|/|\Delta_0|$ & Predicted $\gamma^k$ ($\gamma=0.95$) \\
\midrule
0 & 1.00 & 1.00 \\
1 & 0.81 & 0.95 \\
5 & 0.60 & 0.77 \\
10 & 0.50 & 0.60 \\
\bottomrule
\end{tabular}
\caption{Ablation decay versus the closed-form prediction, the same $\gamma$ fitted independently in Section~\ref{sec:results}.2.}
\label{tab:decay}
\end{table}

The two columns track closely, a falsifiable link between the closed-form account and the causal edit: the edit decays at approximately the time constant the probe itself reads, not some unrelated rate.

A separate measure, whether the agent's routing decision flips, shows the same pattern (0.817 versus $0.247\pm0.019$ null average, 100th percentile, $z\approx+26$).
As a further, externally-grounded check, confusion also tracks a known latent-attractor bias in RSSM world models \cite{berger2026} on held-out cartpole states, correlating with imagined-vs-real drift ($r\approx+0.39$, $p<10^{-146}$) and imagined-reward over-estimation ($r\approx+0.48$, $p<10^{-229}$).

\begin{figure}[H]
\centering
\includegraphics[width=\linewidth,height=2in,keepaspectratio]{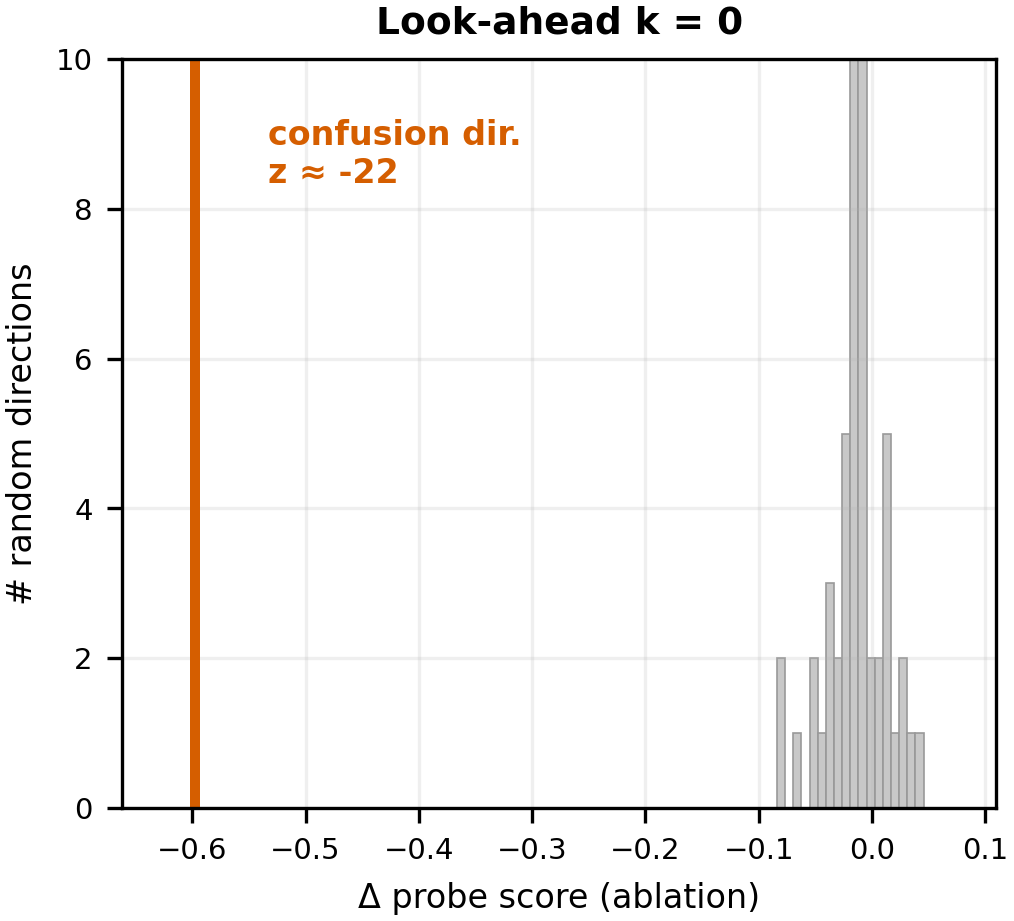}
\caption{Ablation effect against the 50-direction empirical null at look-ahead $k=0$ (the immediate effect). The confusion direction (marked, $\Delta_0=-0.586$) sits at the 100th percentile of the null, $z\approx-22$.}
\label{fig:null_k0}
\end{figure}

\textbf{Robustness.} Rotating the edited direction by up to 0.25 radians retains 58\% of the effect rather than losing it abruptly, consistent with a genuine effect.

\textbf{Two unrelated forward-dynamics measures.} We checked how far an imagined rollout drifts from what actually happens next, and how well the decoder reconstructs the next real observation after the edit. Neither measure separates from its own random-direction null ($|z|<1$ in both cases, one lands at the 28th percentile), while the probe readout and the routing decision separate decisively. Confusion reads dynamics-unreliable states without being the causal lever for that unreliability.

\textbf{Real-value substitution.} For a recipient held-out state, a donor real state is matched on the component orthogonal to the confusion direction (mean match cosine 0.86) but has the opposite confusion level; only the recipient's projection onto the confusion direction is replaced with the donor's real value, so $h_t$ is never edited synthetically. This gives $\Delta_{\text{probe}}=-0.761$ (versus $-0.586$ for the synthetic edit) and a routing-flip rate of 0.868 (versus 0.807), both at the extreme of the null, agreeing with the synthetic result via a structurally different method. The same substitution at later look-ahead gives $\Delta_{\text{probe}}=-0.699,-0.602,-0.534$ at $k=1,5,10$, the same graceful decay pattern as the synthetic edit.

\textbf{Cross-seed replication.} Repeating the protocol on 5 independently trained models, the probe-decay effect sits at the extreme of each model's own null on all 5 ($z$ ranging $-3.9$ to $-8.0$, mean $-0.385\pm0.118$); the routing-flip effect does so on 3 of 5, likely reflecting baseline variability across models; next-step prediction error does not separate from its null on any, matching the single-model result.

\textbf{The boundary's gate dependence.} Forcing the update gate across the same 0.5--0.99 range used for the geometry test (Section~\ref{sec:results}.3):

\begin{table}[H]
\centering
\footnotesize
\setlength{\tabcolsep}{4pt}
\begin{tabular}{cccc}
\toprule
Gate $z$ & Full probe & $\|h\|$ only & $1-z$ \\
\midrule
0.50 & 1.000 & 0.971 & 0.50 \\
0.70 & 1.000 & 0.888 & 0.30 \\
0.90 & 1.000 & 0.622 & 0.10 \\
0.99 & 1.000 & 0.508 & 0.01 \\
\bottomrule
\end{tabular}
\caption{Forced gate value vs.\ AUROC of the full probe and the magnitude-only ($\|h\|$) separator, against the gate's overwrite fraction $1-z$. The magnitude-only separator collapses toward chance as overwriting vanishes; the full probe does not.}
\label{tab:gate}
\end{table}

The magnitude separator tracks the $(1-z)$ overwrite fraction at $r=+0.97$, while the full linear probe's AUROC stays at 1.0 across the whole range. Gate-driven overwriting explains the magnitude effect's size, not why the classes are separable at all, since separability survives even minimal overwriting.

\section{Why Set C Inverts on Pendulum}
\label{app:crossenv}

Pendulum's per-seed Set C scores across 4 runs: 0.322, 0.395, 0.496, 0.371 (mean $0.396\pm0.063$, all four below 0.5). Reacher's per-seed within-task control scores: 0.578, 0.565, 0.723, 0.751 (mean $0.654\pm0.084$), more variable and further above chance than the single run initially suggested.

\textbf{Ruling out the kinematic-hardness hypothesis.} We first checked whether pendulum's inversion was driven by dynamically hard instants, such as near the top of the swing, corrupting the reconstruction-based label.

\begin{table}[H]
\centering
\small
\begin{tabular}{lccl}
\toprule
Task & $r(\mathrm{recon},C_t)$ & Best kinematic $r$ & Dominant driver \\
\midrule
Cartpole & $+0.40$ & $+0.50$ (pole $\omega$) & kinematic \\
Reacher & $+0.40$ & $+0.33$ (velocity) & $C_t$ \\
Pendulum & $+0.33$ & $+0.27$ (upright) & $C_t$ \\
\bottomrule
\end{tabular}
\caption{Reconstruction error correlates more with $C_t$ than with the best available kinematic feature on the two tasks where it matters; cartpole is the one task where kinematics dominate recon, and its Set C test works fine regardless.}
\label{tab:kinematic}
\end{table}

This rules out the kinematic-hardness hypothesis: pendulum's reconstruction error correlates \emph{more} with $C_t$ than with any single kinematic feature, and cartpole, the task where kinematics matter most for reconstruction error, has a working Set C anyway.

\textbf{The actual mechanism.} The explanation is the within-bin correlation between reconstruction error and $C_t$: $+0.39$ on cartpole, $-0.09$ on reacher, $-0.12$ on pendulum, matching the direction of the Set C result on each task (0.72, 0.62, 0.32). Relabelling pendulum's Set C directly by $C_t$ instead of reconstruction error only recovers it to chance (0.53), because at matched KL, KL already explains most of $C_t$ on pendulum. The sign of the within-bin correlation predicts the direction of a possible failure in a reconstruction-based test; how much KL already explains predicts its magnitude.

\begin{figure}[H]
\centering
\includegraphics[width=0.85\linewidth,height=2in,keepaspectratio]{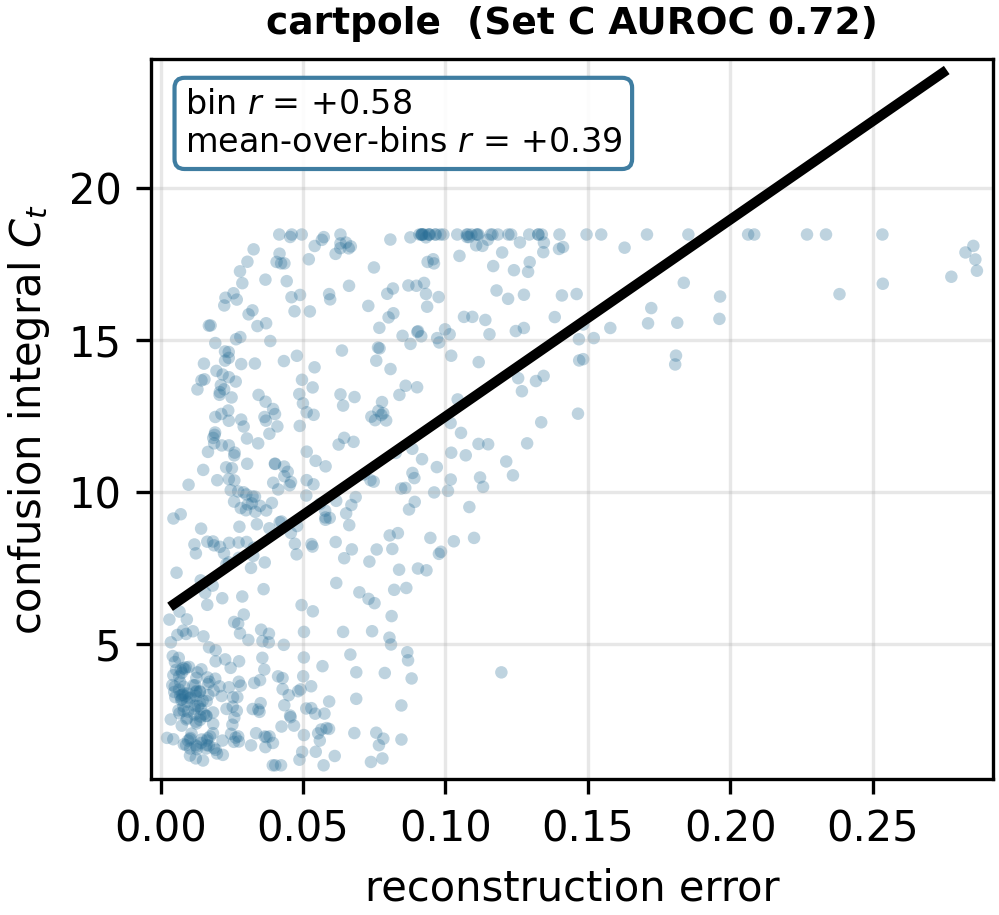}
\caption{Cartpole: reconstruction error against $C_t$ within a single KL bin. The within-bin trend is positive, the recon-based Set C label agrees with confusion.}
\label{fig:setc_mech_cartpole}
\end{figure}

\begin{figure}[H]
\centering
\includegraphics[width=0.85\linewidth,height=2in,keepaspectratio]{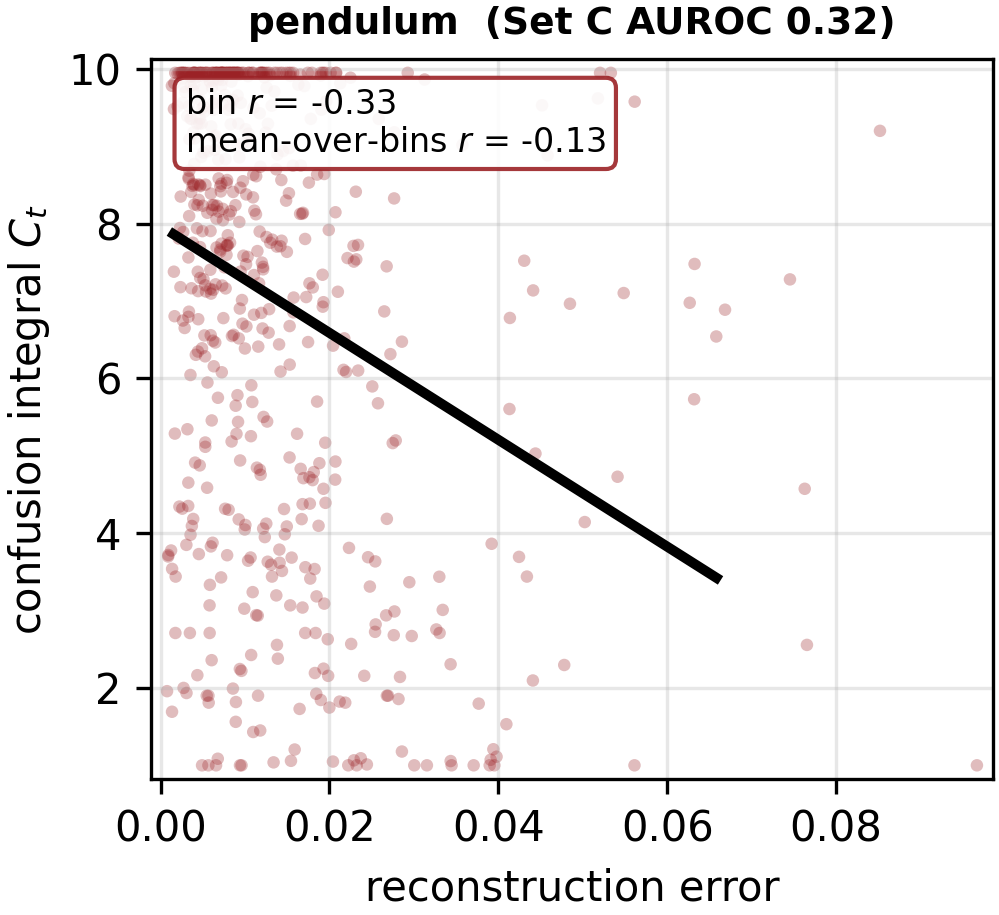}
\caption{Pendulum: reconstruction error against $C_t$ within a single KL bin. The within-bin trend is negative, the sign flip that inverts Set C despite pendulum having the strongest $C_t$ encoding of the three tasks.}
\label{fig:setc_mech_pendulum}
\end{figure}

\section{Scale Check}
\label{app:scale}

Doubling GRU width (to 512 dimensions, one run), a lighter causal test (single random-direction control rather than 50, on 320 held-out states):

\begin{table}[H]
\centering
\small
\begin{tabular}{lcc}
\toprule
Quantity & deter=256 (XS) & deter=512 \\
\midrule
Null-space angle & $\sim 88^\circ$ & $89.4^\circ$ \\
Variance in top-10 PCs & $\sim 0.5\%$ & $0.13\%$ \\
Ablation $\Delta$ (confusion, $k=0$) & $-0.586$ & $-0.287$ \\
Ablation $\Delta$ (random, $k=0$) & --- & $-0.050$ \\
\bottomrule
\end{tabular}
\caption{Both load-bearing findings hold at double the width: the direction stays near-orthogonal to the top PCs, and ablating it collapses the probe far beyond random.}
\label{tab:scale}
\end{table}

This is one additional data point, partial evidence the findings are not XS-specific, not a resolution of the full XL-scale question.

\section{Two Negative Results}
\label{app:negative}

Down-weighting imagined-rollout value estimates by the probe output makes them worse, not better ($\Delta r=-0.526$ vs.\ no weighting); a continuous version of $C_t$ instead of the binary output is worse still ($\Delta r=-0.576$), ruling out binary labelling as the cause. Stopping an imagined rollout early once confusion crosses a threshold (90th percentile, truncating 24\% of rollouts at a mean stop step of 4) rather than down-weighting throughout shows a small positive effect on one split ($\Delta r=+0.009$) that disappears under a stability check across 5 samples (mean $\Delta r=+0.005\pm0.009$, sign unstable). Both approaches agree the signal is useful for detecting problems in imagined rollouts, not for correcting the value estimates computed from them, though early stopping at least does not make things worse.

\section{Routing: Probe-A Versus a $C_t$-Direct Router and a KL-Only Baseline}
\label{app:routing}

Section~\ref{sec:results}.5's Set C inversion on pendulum raises an obvious question: if the recon-based Set C label is what inverts, would a router built directly on $C_t$, bypassing that label entirely, do better than the Probe-A router where Set C misleads? A further question, raised by review, is whether the probe adds anything over simply thresholding the world model's own KL signal. We built both to test directly. All routers are evaluated at a 30\% query budget against the same reconstruction-error baseline, on all three tasks. The KL-only router uses $\mathrm{KL}_{t-1}$, the most recent KL available before the routing decision at step $t$, the same information regime Probe-A operates in; scoring on same-step $\mathrm{KL}_t$ is a tautology against this paper's own event definition (top-25\% KL) and is not reported as a baseline.

\begin{table}[H]
\centering
\footnotesize
\setlength{\tabcolsep}{4pt}
\begin{tabular}{lcccc}
\toprule
Task & Probe-A & $C_t$-direct & KL (prior) & vs.\ recon \\
\midrule
Cartpole & 0.682 & 0.664 & 0.721 & $+0.040$ \\
Reacher & 0.511 & 0.411 & 0.259 & $-0.077$ \\
Pendulum & 0.778 & 0.744 & 0.808 & $+0.297$ \\
\bottomrule
\end{tabular}
\caption{Recall at a 30\% query budget, single held-out split ($N=40{,}000$/task). "KL (prior)" is the KL-only router using $\mathrm{KL}_{t-1}$; "vs.\ recon" is Probe-A's gap against the reconstruction-error oracle.}
\label{tab:routing_full}
\end{table}

\begin{table}[H]
\centering
\footnotesize
\setlength{\tabcolsep}{2.5pt}
\begin{tabular}{lccc}
\toprule
Task (seeds) & Probe-A & KL (prior) & $\Delta$ recall \\
\midrule
Cartpole (5) & $0.626\pm0.047$ & $0.632\pm0.068$ & $-0.006\pm0.025$ \\
Reacher (3) & $0.539\pm0.013$ & $0.267\pm0.007$ & $\mathbf{+0.273\pm0.020}$ \\
Pendulum (3) & $0.729\pm0.017$ & $0.760\pm0.023$ & $-0.030\pm0.006$ \\
\bottomrule
\end{tabular}
\caption{The probe's advantage (mean $\pm$ sd) over a KL threshold, repeated across every existing multi-seed collection. Only reacher shows a reliable, sign-stable win.}
\label{tab:routing_kl_delta}
\end{table}

The dissociation predicted by Section~\ref{sec:results}.5 does not hold for the $C_t$-direct router: it tracks Probe-A closely on all three tasks rather than rescuing reacher or improving on pendulum, and is consistently slightly worse (recall-vs-budget AUC lower by 0.01--0.03 at every task). The KL-only comparison is less kind to the probe. On pendulum, a zero-parameter threshold on last-step KL reliably beats the trained probe (0.808 vs.\ 0.778, 3 of 3 seeds); the paper's pendulum routing gain over reconstruction error ($+0.30$, Table~\ref{tab:routing}) is real, but it is KL redundancy doing the work, not the probe, consistent with Appendix~\ref{app:crossenv}'s finding that KL already explains most of $C_t$ on pendulum. On cartpole there is no reliable probe advantage over KL ($-0.006\pm0.025$, sign unstable across 5 seeds). Reacher is the one task where the probe adds a real, replicated advantage over KL ($+0.273\pm0.020$, 3 of 3 seeds), and reacher is also where $C_t$'s closed-form fit is weakest ($R^2=0.26$) and its memory shortest: where confusion is least reducible to raw KL is where the learned probe earns its advantage. The binding constraint on reacher's routing loss against reconstruction error is the task's dynamics; the binding constraint on the probe's value over KL is exactly the opposite pattern.



\bibliographystyle{named}
\bibliography{ijcai26}

\end{document}